\documentclass[conference,a4paper]{ieeetran}

\usepackage{longtable,flushend}
\usepackage{graphicx}
\usepackage{listings}
\usepackage{tabulary}
\usepackage{csquotes}
\usepackage{footnote}
\makesavenoteenv{tabular}
\makesavenoteenv{table}
\usepackage[]{subfigure}
\usepackage{fancyheadings}
\title{Auto-Detection of Safety Issues in Baby Products}
\author{Graham Bleaney$^1$, Matthew Kuzyk$^1$, Julian Man$^1$, Hossein Mayanloo$^1$, H.R.Tizhoosh$^2$\\
$^1$Systems Design Engineering, University of Waterloo, Waterloo, ON, Canada\\
$^2$KIMIA Lab, University of Waterloo, ON, Canada
}

\begin{document}
\maketitle
\begin{abstract}
Every year, thousands of people receive consumer product related injuries. Research indicates that online customer reviews can be processed to autonomously identify product safety issues. Early identification of safety issues can lead to earlier recalls, and thus fewer injuries and deaths. A dataset of product reviews from Amazon.com was compiled, along with \emph{SaferProducts.gov} complaints and recall descriptions from the Consumer Product Safety Commission (CPSC) and European Commission Rapid Alert system. A system was built to clean the collected text and to extract relevant features. Dimensionality reduction was performed by computing feature relevance through a Random Forest and discarding features with low information gain. Various classifiers were analyzed, including Logistic Regression, SVMs, Na{\"i}ve-Bayes, Random Forests, and an Ensemble classifier. Experimentation with various features and classifier combinations resulted in a logistic regression model with 66\% precision in the top 50 reviews surfaced. This classifier outperforms all benchmarks set by related literature and consumer product safety professionals.
\end{abstract}


\normalfont 
\section{Introduction}
\label{sec:intro}

Thousands of people are injured or killed from consumer product related injuries every year. Between October 1, 2011 and September 30, 2012, the United States Consumer Product Safety Commission (CPSC) recorded over 3,800 deaths that involved consumer products \cite{cpsc2015}. In order to minimize damage to the public, governmental regulatory agencies such as the CPSC are mandated to ensure the safety of consumer products by enforcing product recalls \cite{CPSCcontact}. These organizations particularly target products that are designed for vulnerable populations in their investigations, such as consumer products designed for young children and the elderly \cite{blasiusinterview}.

Despite the efforts of these agencies, product safety issues still occur. One prominent example is the Samsung Galaxy Note7 phone, which was recalled for catching fire \cite{note7recall}. When consumers encounter issues like these, they sometimes voice their concerns about a product in online reviews. In some cases, the complaints are serious enough to warrant an investigation, and possibly a recall. For example, customers were complaining about their Galaxy Note7 phones catching fire as early as August 24th, 2016 \cite{note7pic}, but Samsung only started working with the CPSC on September 9th, 2016 \cite{samsungworking}, and the product was not recalled until September 15th, 2016 \cite{note7recall}. Consumers are exposed to preventable risk in the time between the first customer complaint and the first regulatory action. Earlier detection of product safety issues, using a system such as the one presented in this paper, could reduce consumer exposure to risk and help organizations like the CPSC fulfill their mandate.

Currently, the CPSC detects product safety issues through consumer and manufacturer reports. Consumer reports are collected from a phone hotline and through their online ``SaferProducts.gov" website \cite{cpschotline}. The CPSC also monitors product-related hospital visits through the National Electronic Injury Surveillance System (NEISS) \cite{cpsc2015}. Finally, the CPSC also has a small internet surveillance team that does manual surveillance \cite{cpscinterview1}. The manual nature of their work means that many of these customer complaints go unnoticed. A solution that automatically parses reviews and identifies those that mention a safety issue could lead to earlier investigations for dangerous products. Conversations with the Director of Field Investigations at the CPSC, and an Internet Investigative Analyst at the CPSC, validated the need for such a solution; such a solution would be ``immediately useful" \cite{blasiusinterview}, especially due to their limited manpower. This limited manpower meant that such a solution would also have to be very precise, with one investigator stating they could not tolerate more than 50\% of the safety issues identified being false positives \cite{cpscinterview2}.

Domain literature suggests that it is possible to build a system to automatically identify product safety issues in online reviews, using machine learning and natural language processing. A project by the University of Washington data science team was able to predict food recalls with 45\% precision by classifying AmazonFresh reviews using Term Frequency-Inverse Document Frequency (TF-IDF) features and a Support Vector Machine (SVM) classifier \cite{washingtonstudy}. Another study was able to predict 50\% of Toyota Recalls from posts in the online ToyotaNation forums by using a ``Smoke Word" list and the k-Nearest Neighbor (kNN) classifier \cite{toyotapaper}. In the consumer product space, one study was able to achieve 39\% precision in the top 400 reviews when classifying safety issues in online toy reviews \cite{toystudy} using a ``Smoke Word" list technique developed by Abrahams et. al \cite{abrahams_vehicle_2012}. However, the precision of this solution does not meet the 50\% precision benchmark set by the CPSC agents. 

A major limitation of much of the current literature is the use of ``Smoke Words" to identify safety issues. Using the presence of single words as features removes the broader context of sentences, which could aid in safety issue detection. For example, the word ``screaming" might appear in a review describing a child ``screaming in pain" or ``screaming with delight".

The work presented in this paper is both novel and necessary for the following reasons. First, the study explores a new application area, baby products, which has not been addressed by any previous works. Second, this paper utilizes many machine learning and data collection techniques that have not been previously used to classify product safety issues. Finally, this study establishes the requirements for a practical solution that could be used by product safety organizations such as the CPSC, and surveys machine learning techniques to identify those that can best meet the organizations' requirements. 

The objective of this study was to build a system that automatically detects consumer product safety issues from online reviews and meets the benchmarks outlined by the CPSC. As a starting point, consumer reviews in the baby product category were targeted, due to their status as a vulnerable population. Various machine learning tools and techniques were tested in order to achieve this objective.

\section{Materials}
\label{sec:materials}

The primary dataset used for review classification was the Amazon review corpus of 142 million reviews provided by Dr. McAuley of UCSD \cite{amazoncorpus}. From this corpus, approximately 7,000,000 reviews in the ``Baby" product category were available for experimentation. From these reviews, the data needed to be labelled for use in supervised machine learning. In cases where it was not clear whether a review counted as mentioning a ``safety issue", the rules in Table \ref{tab:labeling_rules} were used.

\begin{table*}[h]
\vspace{0.1in}
    \centering
    \caption{Rules for manual labelling}
%    \begin{tabular}{p{0.25\textwidth} p{0.19\textwidth}}
    \begin{tabular}{p{0.85\textwidth} p{0.10\textwidth}}
        \hline
             \textbf{Issue Mentioned in Review} & \textbf{Safety Issue?}  \\
        \hline
             Person harmed during correct use of the product & \textbf{Yes} \\
        \hline
             Person harmed during incorrect use of the product & \textbf{Yes} \footnote{CPSC suggests firms test for and correct issues arising from foreseeable misuse \cite{cpscmisuse}} \\
        \hline
             Harm could have occurred, but was avoided through an action by the user & \textbf{Yes} \\
        \hline
             Different product (i.e., not the product the review is associated with) has a safety issue & \textbf{Yes} \footnote{It is important to highlight potential safety issues, even if tied to a wrong product causing more follow up} \\
        \hline
             Potential harm is suggested, with no evidence  (e.g., ``This is made in China, so it's probably full of toxic chemicals") & \textbf{No} \\
        \hline
    \end{tabular}
    \label{tab:labeling_rules}
\end{table*}

Given the rarity of reviews mentioning product safety issues relative to the size of the whole review corpus, the search space for reviews mentioning safety issues had to be reduced. This was done by combining the CPSC's list of recalled products with the Amazon review dataset, using Universal Product Codes (UPCs). These reviews were filtered to ensure that they were written before the official recall date, to imitate the data that would be available to the system in real-world usage. In total, 2,285 reviews were labeled.

Throughout the testing process, classifiers were tested on unlabelled Amazon reviews (discussed in Sect. \ref{sec:validation}), and the results were labeled. Combining these reviews with the previous labelling, led to a total of 3,773 labelled reviews available for the final training run presented in this paper, 424 of which mentioned product safety issues.

Manual analysis showed that the labelled reviews mentioned only a subset of known safety issues, and skewed toward low severity ones. For these reasons, the labelled data was augmented with other datasets, including SaferProducts.gov complaints, CPSC recall descriptions, and European Commission Rapid Alert data. All of the datasets used are shown in Table \ref{tab:datasets}. Since the reports from these additional datasets explicitly mention harm, they were considered labelled as ``Mentions Safety Issue". Each dataset was downsampled to 3,333 complaints due to computational constraints of the hardware available. These were sampled randomly to avoid bias.
   
\begin{table*}[h]
    \centering
    \caption{Explanation of datasets used in final evaluation}
    \begin{tabular}{ p{0.23\textwidth}    p{0.25\textwidth}    p{0.44\textwidth}}
        \hline
            \textbf{Name} &
           \textbf{Number of Complaints} & \textbf{Description}
           \\
        \hline
            Amazon Reviews & 3,773 labelled/used, 7,155,102 unlabelled.
            & Reviews of products on Amazon. \\
        \hline
            SaferProducts.gov
            & 30,386 labelled, 3,333 used
            & Reports submitted to the CPSC indicating harm caused by a consumer product. \\
        \hline 
            CPSC Hazard Descriptions
            & 14,835 labelled, 3,333 used
            & Descriptions that accompany CPSC recalls. \\
        \hline 
            European Commission Rapid Alert 
            & 51,272 labelled, 3,333 used
            & Descriptions that accompany EU recalls. \\
        \hline        
    \end{tabular}
    
    \label{tab:datasets}
\end{table*}
Additional materials included ``Smoke Word" lists: lists of words that could indicate danger if used in a review. The ``Smoke Word" list used was compiled by combining a ``Smoke Word" list provided by analysts at the CPSC \cite{cpscinterview1}, a ``Smoke Word" list generated by Abrahams et al. in their analysis of children's toy safety issues \cite{toystudy}, and a custom ``Smoke Word" list generated from the labelled reviews dataset by computing TF-IDF (see Sect. \ref{sec:features}) feature importance. This generated ``Smoke Word" list was reviewed, and any words that did not seem to indicate an issue were removed.

\section{Methods}
% Julian
% \begin{itemize}
%     \item Briefly describe background/motivation for:
%     \item TF-IDF (incl. bigrams)
%     \item Doc2Vec
%     \item Star Rating
%     \item Wordlists (CPSC, Abrams, Abrams/Neiss)
%     \item Cleaning: Stemming, non-english removal, etc.
%     \item Mention the feature important shit we do to go from 160K -> 1K features (approx numbers)
% \end{itemize}

\begin{figure*}[h]
    \centering
    \includegraphics[width=\textwidth,keepaspectratio]{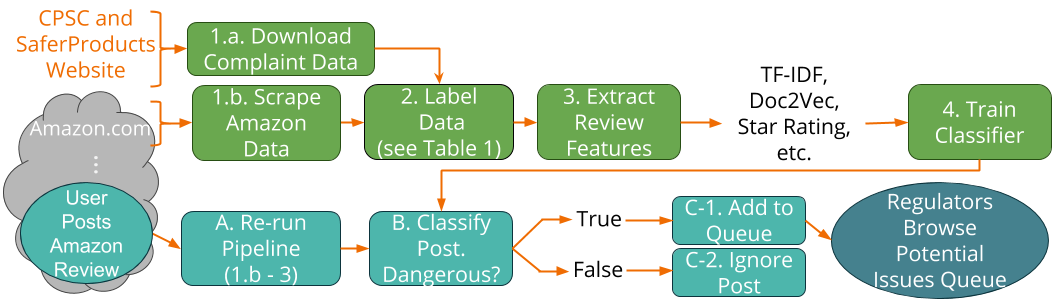}
    \caption[]{System diagram for the product safety issue detection system.}
    \label{fig:system_diagram}
\end{figure*}

A system diagram for the product safety issue detection system is shown in Fig. \ref{fig:system_diagram}. The system includes a data collection (scraping and downloading) component, a machine learning component, and a user interface component; this paper focuses mainly on the machine learning component. The machine learning system was comprised of three sub-components: a text preprocessing component, a feature extraction component, and a classification component. These components can be run with no human interaction, allowing near-instantaneous identification of reviews mentioning product safety issues, reducing the time it takes organizations like the CPSC to identify product safety issues.

\subsection{Text Preprocessing}
Prior to extracting features, the review text was cleaned by removing non-English words, converting to lower case, and stemming. Non-English word removal was important for removing proper nouns, such as product brand names. While brand names may sometimes provide context for product quality, not removing brand names was found to bias the classifier against large brands that previously had one particularly public recall. Stemming is a heuristic process to convert words to a base form. For example, stemming might convert ``safer" and ``safely" to the same base stem ``safe", to help classifiers treat similar words identically. The snowball stemmer in the NLTK library was used for stemming.

\subsection{Feature Extraction}
\label{sec:features}

Following the text cleaning process, feature vectors were generated from each review. The results of each feature extraction technique were concatenated to form a single feature vector per review. TF-IDF, Doc2Vec, star rating, and ``Smoke Word" count features were all used. 

TF-IDF converts text to a vector of word counts weighted by word rarity. This representation helps classifiers identify words that only appear in a small number of documents. Bigram (pairs of words) counts, in addition to unigrams, were computed because they help handle negation \cite{pang_thumbs_2002}.

Doc2Vec is an unsupervised neural network model, based on the word2vec model, for converting text into a vector representation that retains semantic information \cite{lau_doc2vec_2016}. A custom Doc2Vec model was trained on a subset of the Amazon review corpus, and used to extract Doc2Vec features from text. The subset consisted of 500,000 reviews randomly sampled from the corpus.

Star ratings are Amazon's user generated numerical product ratings. They were included as a feature to proxy the user's sentiment. For the non-Amazon datasets (e.g. recalls and consumer complaints), the star rating was assumed to be one star, due to the negative sentiment of the data.

``Smoke Word" lists are curated lists of words indicative of safety issues. These ``Smoke Word" lists were acquired from multiple sources, as discussed in Sect. \ref{sec:materials}. ``Smoke Word" count was computed from these ``Smoke Word" lists by counting total occurrences of all smoke words that appeared in a review.

Feature extraction yielded 256,000 features, which was reduced to 2,400 features using a Random Forest to compute feature importance. Feature information gain was calculated as part of the Random Forest training process, with low gain features being discarded.

\subsection{Classification}
\label{sec:class}
% TODO: Cut explanations if desperate

Several commonly used classifiers were tested: SVM, Logistic Regression, Na{\"i}ve-Bayes, Random Forest, k-Nearest Neighbor (kNN), and an Ensemble classifier.

SVM, Logistic Regression, and Na{\"i}ve-Bayes classifiers were used because they were effective for text classification in literature \cite{khorsheed_classifiers_2013}. SVMs find the linear separating hyperplane with maximum distance from the closest points of each class. When the data is not linearly separable, kernels are used to project data to a higher dimension where the data is linearly separable \cite{suykens_svm_1999}. In this study, linear kernels were used. Logistic Regression fits the data to a logistic distribution and results in a linear separating hyperplane \cite{ng2002discriminative}. L2 regularization with parameter $\lambda=0.001$ was used. Na{\"i}ve-Bayes classifiers use Bayes' conditional probability formula to predict the most probable class for a given data point \cite{ng2002discriminative}. The classifier makes the na{\"i}ve assumption that features are conditionally independent. 

Random Forests combine the predictions of a large number of small decision trees, which are trained on separate subsets of features and training examples \cite{liaw2002randomforest}. Each random forest had 10 trees, each trained with the square root of the total number of features. kNN is a simple classifier in which the $k$ closest training examples vote on the class of a new data point. kNN has been used in literature for text classification \cite{toyotapaper}. The $k=5$ nearest neighbors were used in this study. Ensemble classifiers aggregate the results of several other classifiers \cite{ruta2005classifier}. In this study, the Ensemble classifier aggregated the results of the other classifiers by averaging their model scores.

Classifiers output a model score for each review, between zero and one. A high model score indicated a high probability that the review ``Mentions Safety Issue". The threshold for this classification was chosen by finding the threshold with the maximum F1 score in the training dataset. F1 score is a measure of classifier performance, which is outlined in Sect. \ref{sec:validation}. 

\subsection{Validation}
\label{sec:validation}

Classifier performance was compared using peak F1 score, and classifier output on the 50 lowest model score and 50 highest model score reviews.

F1 score is the harmonic average between the precision (percentage of reviews classified as ``Mentions Safety Issue" that actually mentioned safety issues) and recall (percentage of reviews mentioning safety issues that the classifier was able to correctly identify). F1 score is a common measure of classifier performance \cite{rijsbergen_information_1979}. F1 score was measured for only reviews from the Amazon dataset. Performance was validated using 5-fold cross validation, a technique commonly used in literature \cite{abrahamsdishwasher}\cite{baselinesbigrams}. This involved dividing the data into five partitions (or folds), running five training cycles while holding out a different partition each time for validation, and training on the other partitions.

Classifier behavior on the 50 lowest model score and 50 highest model score Amazon reviews was examined using a confusion matrix, showing occurrences of their predicted class versus their actual class. In this case, confusion matrices show true positive, false positive, true negative, and false negative counts, where a ``positive" example is a review indicating a safety issue. Examining the confusion matrix of the highest and lowest model score reviews is a technique used in literature \cite{toystudy}. Measuring precision on the top 50 reviews, simulated the real world usage of the classifier; investigators are likely to only examine very high model score reviews, given the volume of reviews on the internet \cite{cpscinterview2}.

In this study, the classifiers were run on 100,000 previously unseen reviews. The top and bottom reviews were manually labelled (see Sect. \ref{sec:materials} for rules). Reviews were labelled by two raters, in order to validate the labels by calculating inter-rater agreement. Fleiss' kappa statistic was found to be $\kappa=0.713$, which indicated ``substantial agreement" \cite{landis1977measurement}. 

\section{Results}
\label{sec:results}

The classifiers outlined in Subsection \ref{sec:class} were trained on the combination of labelled baby product reviews, recall descriptions, and consumer complaints that were described in Sect. \ref{sec:materials}. The results are presented in Table \ref{tab:superclassifier}, with the best results highlighted in bold. Each row contains confusion matrix data on true positives (TP), false positives (FP), true negatives (TN), and false negatives (FN). The Ensemble classifier had the highest F1 at 0.491, SVM had the highest precision at 53.4\%, and Logistic Regression had the highest recall at 70.8\%.

\begin{table}
\vspace{0.1in}
    \centering
        \caption{Classifier results for training on Amazon reviews, recall descriptions, and consumer complaints, and tested on only Amazon reviews. Best results are highlighted in \textbf{boldface}}
    \begin{tabular}{l r r r r r r r}
    \hline
        Classifier & F1 & Precision & Recall & TP & FP & TN & FN \\
    \hline
        kNN & 0.430 & 0.347 & 0.590 & 191 & 372 & 2220 & 133 \\
        \textbf{Logistic Regression} & 0.457 & 0.339 & \textbf{0.708} & 229 & 452 & 2140 & 95 \\
        Na{\"i}ve-Bayes & 0.387 & 0.469 & 0.334 & 108 & 129 & 2463 & 216 \\
        Random Forest & 0.442 & 0.387 & 0.527 & 171 & 279 & 2313 & 153 \\
        \textbf{SVM} & 0.451 & \textbf{0.524} & 0.398 & 129 & 117 & 2475 & 195 \\
        \textbf{Ensemble} & \textbf{0.491} & 0.491 & 0.491 & 159 & 166 & 2426 & 165 \\
    \hline

    \end{tabular}
    \label{tab:superclassifier}
\end{table}

A random sample of 100,000 unlabeled Amazon reviews were classified by each trained classifier, and the reviews with the highest and lowest scores were considered to be labelled as ``Mentions Safety Issue" and ``Does Not Mention Safety Issue" respectively. The computation time taken to classify the reviews was negligible. These reviews were then manually labelled to validate the classifier-assigned labels.

Table \ref{tab:manual_label_results} shows classifier precision, calculated in terms of the positive --- ``Mentions Safety Issue" --- class. Recall was not calculated, because there was no data on the total number of safety issues in the 100,000 unlabelled reviews. A few reviews in the random sample were not in English, and were removed from the analysis as a result.

The best performing classifier was the Logistic Regression classifier, with 66\% precision. This result surpassed all of the benchmarks defined, which will be discussed further in Sect. \ref{sec:discussion}. To determine the significance of these results, 50 randomly sampled 1-star reviews for baby products were manually labelled. 16\% of these 1-star reviews mentioned a safety issue. A chi-squared test demonstrated with $p<0.0001$ that the Logistic Regression classifier results were a statistically significant improvement over randomly selecting 1-star reviews. 

\begin{table}
    \centering
        \caption{ Evaluation of classifiers based on reviews that they assigned the highest and lowest scores. Best results are highlighted in \textbf{boldface}}
    \begin{tabular}{l r r r r r}
    \hline
        Classifier & Precision & TP & FP & TN & FN \\
    \hline
        kNN & 0.163 & 8 & 41 & 50 & 0 \\
        \textbf{Logistic Regression} & \textbf{0.660} & 33 & 17 & 50 & 0 \\
        Na{\"i}ve-Bayes & 0.478 & 22 & 24 & 48 & 2 \\
        Random Forest & 0.260 & 13 & 37 & 50 & 0 \\
        SVM & 0.560 & 28 & 22 & 49 & 1 \\
        Ensemble & 0.490 & 24 & 25 & 50 & 0 \\
    \hline

    \end{tabular}

    \label{tab:manual_label_results}
\end{table}

Below are some examples of reviews that were identified by the Logistic Regression classifier:

\begin{itemize}
    \setlength\itemsep{1em}
    \item ``... my 3 year old opened the window anyways and \textbf{fell out of the second story window}..."
    \item ``...\textbf{YOUR CHILD CAN CHOKE ON THIS TOY!} My 5-month old daughter was playing when one of the legs became \textbf{lodged in her throat}..."
\end{itemize}

The reviews causing false positives and negatives were investigated, and two main reasons were identified: negation and lack of severity. Negation caused reviews such as this one to be mistakenly classified as ``Mentions Safety Issue": ``We hand washed them and air dried them, but \textit{never had mold}". Some false positive reviews only mentioned low severity issues: ``I am really concerned this \textit{odor is toxic}", which did not meat the criteria of being a safety issue (Tab. \ref{tab:labeling_rules})

\section{Discussion and Conclusions}
\label{sec:discussion}

\begin{figure}
\vspace{0.1in}
    \centering
    \textbf{Classifier Performance and Precision Benchmarks}\par\medskip
    \includegraphics[width=0.95\columnwidth,keepaspectratio]{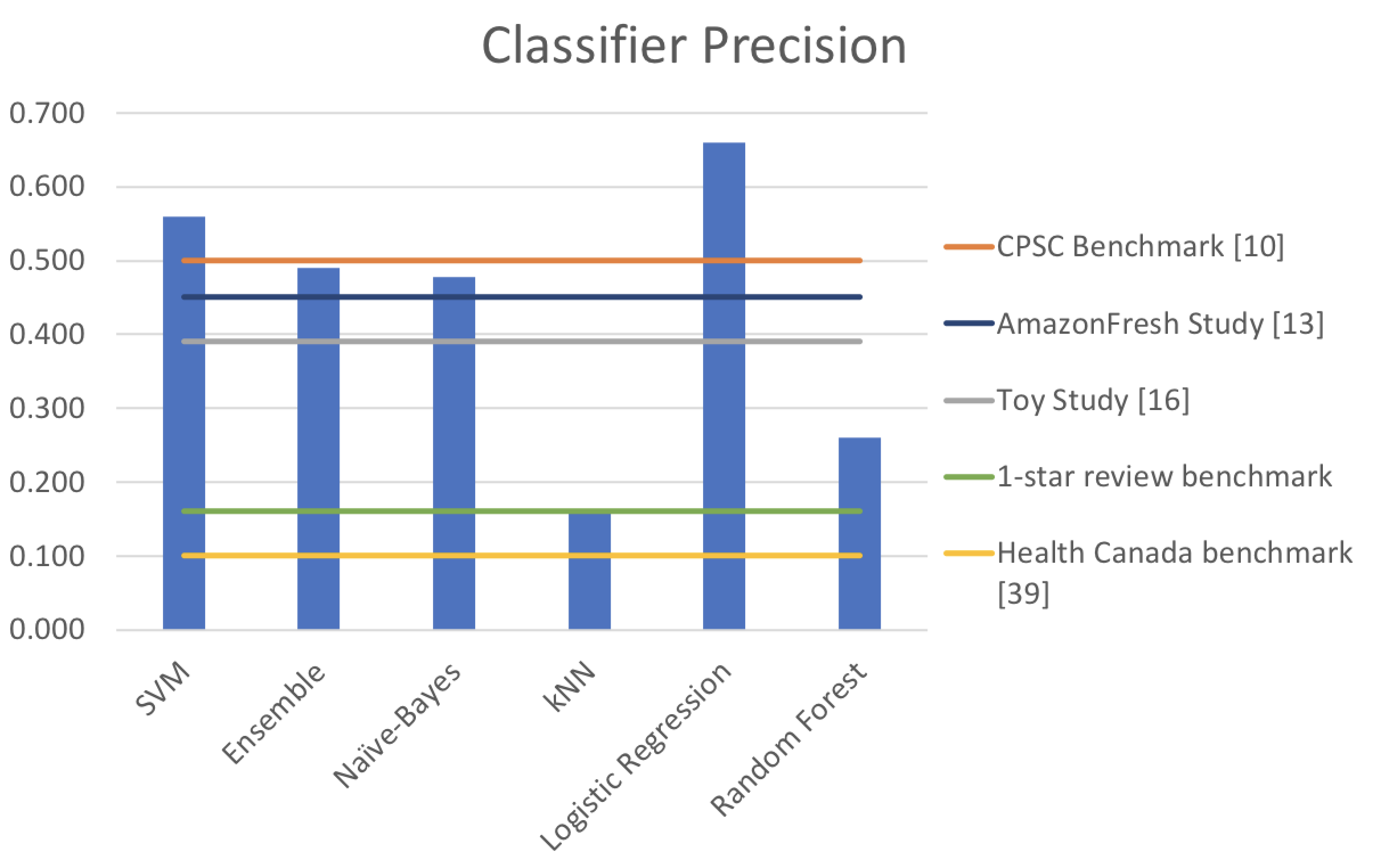}
    \caption[]{Classifier precision and benchmarks set by similar studies, and investigators from government agencies.}
    \label{fig:precision}
\end{figure}

Different classifiers were trained to identify product safety issues mentioned in online baby product reviews. SVMs, Logistic Regression, and Na{\"i}ve-Bayes classifiers performed well, an observation which agrees with reports in literature \cite{khorsheed_classifiers_2013}. Precision benchmarks and classifier precision on the top 50 reviews are summarized in Fig. \ref{fig:precision}. The Logistic Regression classifier had 66\% precision, with the next highest being SVM at 56\% precision. These classifiers outperformed the precision benchmarks set in literature. Winkler et al. used a ``Smoke Word" list and achieved 39\% precision when identifying product safety issues in Amazon reviews of toys \cite{toystudy}. Winkler et al. used a different subset of the same review corpus, and a similar evaluation methodology as this study (measuring precision on the top reviews output by the classifier), and so is most closely comparable to the results found here. Another study predicted food product recalls with 45\% precision on AmazonFresh review data, using an SVM classifier \cite{washingtonstudy}. A key difference with that study is it evaluated results against the entire dataset scored by the classifier, rather than just reviews with the top model scores. The Logistic Regression and SVM classifiers also exceeded the performance benchmarks set by users: the CPSC required a minimum precision of 50\% \cite{cpscinterview1}, and a former Health Canada investigator required a precision of 10\% \cite{christineinterview1}. Finally, the performance exceeded the 16\% precision which was achieved in Sec \ref{sec:results} by assuming all one star reviews mention product safety issues.

% This limitation can be overcome using knowledge transfer \cite{neural_net_knowledge_transfer}. This is where a neural network is trained for a task for which a large quantity of data is available, and repurposed for the low data volume task. 

While the results outperformed all benchmarks set, there were still many false positives due to issues such as negation. ``Negation'' can be addressed by using larger n-grams. The current approach uses bigrams, which only capture negation within a one word radius \cite{qu2010bag}. However, using larger n-grams would result in an exponential increase in the dimension of the feature space \cite{qu2010bag}. The extremely sparse feature space may result in classifier performance issues. An alternative approach is using sentiment analysis on a per-sentence level, allowing classifiers to discount the presence of danger words in overall positive sentences.

To address reviews with low severity issues being classified as dangerous, results may be improved by adding severity levels to the ``Mentions Safety Issue" label. Additionally, experts from agencies such as the CPSC could be used to label the data based on their stricter severity guidelines.

The performance of the classifiers were limited by the quantity of labelled reviews. Labelling reviews is a time consuming process at scale. Crowdsourcing platforms such as CrowdFlower or Amazon Mechanical Turk can be used to outsource the labelling work. To ensure the quality of the outsourced work, some expert labelled data could be included and contributors with low inter-rater agreement (see Sect. \ref{sec:validation}) could have their labels excluded.

It may be possible to further improve performance using neural networks \cite{conv_neural_net}. Neural networks are extremely powerful classifiers, requiring large quantities of labelled data. Future work may explore the application of neural networks to safety issue detection in online reviews. % Delete the second half of that sentence if you need space

The implemented solution was limited to English baby product reviews. There is reason to believe that it is possible to apply this approach to all product categories. Every dataset used has data across all product categories, so expanding the scope of the system is as simple as repeating the process described in this paper. Other languages can be supported by following the same process with analogous datasets. For example, Health Canada reports all Canadian recalls in French \cite{healthcanadafrench}, which is a good analog for CPSC recall data. Additionally, Amazon operates non-English domains, which could be used for non-English review data.

The system described in this paper is almost ready for practical use. It continuously scrapes Amazon reviews throughout the day, and classifies them. In a real world setting, users of the system would be providing feedback on the reviews the system highlights (i.e., true vs false positive), which could be used to continuously retrain and improve the classifiers.

The system can help reduce consumer product related deaths and injuries by reducing issue detection time, and thus remediation time. The classifier was developed alongside a user interface, which was tested with CPSC and Health Canada investigators, and met their requirements. This work also has applications in industry: companies can monitor online reviews for their products, and quickly identify and rectify safety issues. It is the authors' hope that this work will be used to improve consumer product safety where possible.

\section*{Acknowledgments}
The authors would like to thank Dr. Olga Vechtomova (University of Waterloo, Canada) for her guidance. The authors appreciate the insights on product safety agencies provided by Christine Simpson (former Health Canada Product Safety Officer), Dennis Blasius (CPSC, Director of Field Investigations Division), Michelle Mach and Renee Morelli-Linen (both CPSC Internet Investigative Analysts). The authors would also like to thank Dr. Alan Abrahams for his advice and for providing his smoke word list. Finally, the authors would like to thank Dr. Julian McCauley for providing his corpus of Amazon reviews.

%\printbibliography
\bibliographystyle{IEEEtran}
\bibliography{sources}
\end{document}